\documentclass[11pt]{article}
\usepackage{coling2018}
\usepackage{times}
\usepackage{url}
\usepackage{latexsym}
\usepackage{graphicx}
\usepackage{amsmath}
\usepackage{wrapfig}

\title{The Historical Significance of Textual Distances}

\author{Ted Underwood \\
  School of Information Sciences and Department of English \\
  University of Illinois, Urbana-Champaign \\
  {\tt tunder@illinois.edu} \\}

\date{}
\begin{document}
\maketitle
\begin{abstract}
  Measuring similarity is a basic task in information retrieval, and now often a building-block for more complex arguments about cultural change. But do measures of textual similarity and distance really correspond to evidence about cultural proximity and differentiation? To explore that question empirically, this paper compares textual and social measures of the similarities between genres of English-language fiction. Existing measures of textual similarity (cosine similarity on tf-idf vectors or topic vectors) are also compared to new strategies that strive to anchor textual measurement in a social context.  
  \end{abstract}

\section{Introduction}
\label{intro}

\blfootnote{

    \hspace{-0.65cm}  % space normally used by the marker
    This work is licensed under a Creative Commons 
    Attribution 4.0 International License.
    License details:
    \url{http://creativecommons.org/licenses/by/4.0/}.
}

Computational methods appeal to humanists partly because they promise to shed light on long-standing questions about the pace and direction of historical change. One popular approach to this topic begins by measuring the similarities between documents (or document parts). Researchers then interpret textual similarity as evidence of social continuity, and textual distance as evidence of change.

Distance has been measured in a range of ways. Researchers often topic model their corpus first---hoping, perhaps, to produce a measure of distance fitted to the patterns of a specific corpus. Once texts are translated into topic vectors, the vectors may be compared using cosine similarity or KL divergence. These methods have recently been used to make arguments about literary and political influence ~\cite{jockers:macroanalysis,barron:individuals} and about the pace of change in popular music~\cite{mauch:evolution}.

The key link in these arguments is the premise that measurements of textual (or acoustic) distance can be used as a proxy for human judgments about social difference. The assumption is not baseless: researchers have shown that simple textual models can detect cultural categories like genre~\cite{kessler:automatic,kim:genre}. So we do know that social differences leave a textual trace. But the converse proposition is less well supported: we don't know that all textual differences will be relevant to a given cultural or political question. The rise of contractions in the 18th century, for instance, is not usually considered a revolution in English literature.

To make its historical significance clearer, textual distance could be measured in relation to specific reference points that define a space of socially meaningful variation. This is easy to achieve when researchers are interested in a single category. To measure divergence from a particular paradigm of science fiction, for instance, researchers can just train a model of the genre using novels that 1950s reviewers labeled ``science fiction.'' If the model is accurate, we know it has defined a space of textual differences that mattered in relation to one point of reference.

But in practice, historians are often more interested in changes that transform the reference points themselves. Genres are not stable. A scholar might wonder, for instance, whether fantasy and science fiction are more distinct today than they were before \textit{The Lord of the Rings.} Since questions of this kind involve the relative positions of multiple generic reference points, it is not immediately obvious how they would be solved with a supervised model. Instead, scholars have to fall back on generalized distance measures, hoping that those measures will roughly correspond to socially meaningful differences.

This paper aims to test that assumption. How rough is the ``rough correspondence'' between textual similarity and social proximity? The paper will:

\begin{enumerate}

\item Introduce evidence about the social proximity of genres in nineteenth- and twentieth-century fiction.
\item Use that evidence to ask whether prevailing measures of textual distance correspond to anything outside the text.
\item And finally, introduce new measures of distance that rely on the triangulation of multiple social reference points instead of generalized comparisons between text vectors.

\end{enumerate}

To avoid misunderstanding: the ``distances'' discussed here will be statistical distances, but not often distance metrics in a strict sense~\cite{cha2007comprehensive}. They may not, for instance, be limited to positive values.

\section{Data}
\label{data}

The underlying data for this paper comes from HathiTrust Digital Library, which contains more than 16 million volumes. Using a mixture of metadata and predictive modeling, we identified 210,305 of those volumes as English-language fiction. Over the course of the past century or so, those books were assigned to subject and genre categories by librarians, using a controlled vocabulary maintained by the Library of Congress. But the controlled vocabulary governing categorization has changed enormously over the last century. 

\subsection{Mutability of genre categories}
Computational methods are often tested on a limited context---say, newsgroups in the 1990s---where categories can be treated as stable. But in many domains that interest historians, this is not true.

The categorization of fiction is a good example. This article will measure the textual differences between ``genres.'' But the concept of genre lacks a consensus definition. Linguists like Biber and Conrad ~\shortcite{biberconrad:genre} may describe genres by enumerating the ``expected textual conventions" that define them (144). But sociologists often bracket textual content, in order to understand genres simply as ``sets of artworks classified together on the basis of perceived similarities'' ~\cite{dimaggio:genre}. The social act of classification itself is foregrounded, and the hypothesis that those classifications are founded on real formal characteristics is left as a conjecture.

Both approaches to genre have validity; a researcher needn't make a final decision between them. But it is important to recognize that genre is a multifaceted concept that has been used differently in different communities. This slipperiness only increases as we move back in time. In fact, it has not always been clear that the category of genre should be central to the classification of books. By the late 19th century, ``subject headings'' had become common in libraries~\cite{stone:lcsh}. But it took longer for librarians to start categorizing books by genre. At first, genres were often treated as subjects~\cite{miller:genre}. Instead of describing works of early-20th-century fiction as \textit{examples} of ``Historical fiction'' or ``Love stories,'' for instance, catalogers described them as books \textit{about} ``History'' or ``Man-woman relationships''---as if they were nonfiction. In the past thirty years, catalogers have applied genre labels to fiction more consistently, but books cataloged earlier usually still have subject headings rather than genre labels. Audience designations (``Juvenile fiction'') and forms (``Short stories'') are also still promiscuously mixed with genre categories.

In short, genres are typical of the categories that interest humanists: they are tangled up with time, for reasons that have as much to do with the history of observers as with the history of the object. A fuller history of genre might draw on a wider range of sources---publishers, for instance, and book reviewers. But the underlying challenge would remain the same: different observers have divided the world of fiction in different ways.  We may strongly suspect that novels under the subject heading ``Detectives'' will resemble those later assigned to the genre category ``Mystery fiction.'' But this is an empirical question about the continuity of social practices.

\subsection{Social proximity}
One way to test the affinities of genres is to measure the overlap between categories in the library itself. Books can carry multiple tags for genre and subject; if certain tags tend to be assigned to the same books, we might infer that those pairs of categories are related in practice. A loosely similar approach is adopted in Wu et al.~\shortcite{wu:genre}. Digital libraries provide plenty of overlapping categories, because they often include multiple copies of a book, cataloged in different libraries. After deduplication, the 210,305 volumes we began with boil down to 138,164 distinct titles, each of which might carry tags assigned by several different hands.

To quantify the tendency to overlap, we calculated pointwise mutual information for every combination of categories $a$ and $b$. PMI is often used to measure the strength of collocations. In this context, however, it potentially overlooks the problem that categories may have disjoint chronological distributions. If the subject category ``History'' was only assigned before 1980, and then got replaced by the genre category ``Historical fiction,'' the two categories might rarely coincide, and appear unrelated. 

We addressed this problem in several ways. First, we calculated the probabilities that are components of PMI only \textit{within} a random sample $t$ of volumes selected to have the same chronological distribution as $a \cup b$. We also added a small constant ($0.1$) to the counts of $(a, b)$ for Laplace smoothing, since the distance from a count of zero to a count of one can be very large in sparse data. The net effect of these changes was to increase PMI for chronologically disjoint categories. 
\begin{equation}
pmi(a;b) = log\frac{p(a, b|t)}{p(a|t)p(b|t)}
\end{equation}
Finally, we supplemented empirical evidence with strong priors about pairs of categories where a genre term and a subject term were identical or closely related. (We expect the subject heading ``Horror,'' for instance, to be closely related to the genre category ``Horror.'')

Even after this adjustment, there are reasons to doubt that we can capture all the relations between genres by measuring the intersection of categories in a library. Two genres in themselves very different might happen to hybridize a lot, producing anomalously high PMI. Moreover, collisions between genres are sparse: in 223 of 496 comparisons, there was no overlap at all between $a$ and $b$.

So PMI based on library metadata should not be taken as ground truth about the real relations between genres. No single oracle can be trusted on that question. Rather, we have many different ways to compare genres, and we want to inquire about the degree of correlation between them. The advantage of a measure based on cataloging practices is simply that it doesn't rely directly on the text; it thus gives us a yardstick plausibly independent from the textual measures we want to evaluate.

\subsection{Textual evidence}

While social proximity was calculated using the entire set of 138,164 titles, textual distances were calculated for a smaller sample, balanced so we would have the same number of examples of each category. We selected twenty genre categories and twelve subject headings, for a total of 32 categories. For each category, we randomly drew 100 volumes as a primary sample.

Because genre categories intersect, it often happened that some volumes randomly selected as ``War fiction'' also carried the tag ``Love stories.'' The intersection of genres is a real social fact, and researchers shouldn't ignore it. But because our measure of social proximity was explicitly based on the size of this intersection, we didn't want textual comparisons to be dominated by the same factor: in that case, we might no longer have two independent measures of similarity. To permit comparison between strictly non-overlapping samples, we selected supplementary groups of books tagged (for instance) ``War-Not-Love'' and ``Love-Not-War,'' which could be used to replace books carrying both tags. 

We also selected two completely random samples of fiction to be used as contrast sets in predictive modeling. Adding together all of these samples, we had 6846 volumes of fiction.

Word frequencies for all these books (even for books under copyright) are publicly available from HathiTrust Research Center~\cite{capitanu}. Because works of fiction sometimes begin with nonfiction introductions, we ignored the first 10\% and last 5\% of pages in each volume. A few categories of individually rare tokens were consolidated into a single collective feature: for instance, all Arabic numbers became ``\#arabicnumber.''

\section{Methods}
\label{meth}

We evaluated three different measures of the textual distance between genres, looking in each case at the correlation with social evidence of genre proximity.

\subsection{Tf-idf vectors}

One well-established measure of the distance between documents multiplies the frequency of each term by its inverse document frequency~\cite{sparck:tfidf}. Distance is usually measured by taking the cosine between the vectors thereby constructed. 

We adapted this to make comparisons between genres by summing the term frequencies of all 100 volumes in the primary sample for each genre to produce a collective term-frequency vector. The top 10,000 words, by document frequency, were considered.

\subsection{Topic vectors}

It has recently become more common for researchers to estimate cultural change by topic modeling a corpus and comparing topic vectors~\cite{jockers:macroanalysis,barron:individuals}. This is in part simply a dimension-reduction strategy, but researchers may also hope that topic vectors will produce a more significant measure of distance. However, it is not entirely clear what kind of significance is being maximized. The generative logic of LDA doesn't in itself guarantee that topics will be optimal for any particular discriminative task~\cite{NIPS2008_3599}. Social evidence about genre similarity allows us to ask how much is really added by the topic modeling step.

We used the scikit-learn implementation of Latent Dirichlet Allocation~\cite{scikit-learn}. Removing a standard list of English stopwords gave us a lexicon of 28,443 words. A model with 100 topics was selected. Topic vectors were then constructed in several different ways:

\begin{enumerate}

\item A na\"ive approach simply summed the topic vectors for the 100 volumes in the primary sample for each genre; this was analogous to the construction of tf-idf vectors, although it tended to weight books more evenly (where the tf-idf method had given more weight to longer volumes).

\item To more carefully ensure that our measure of similarity was not defined by the size of the intersection between genres, we also made comparisons between the symmetric differences of genre sets. This required constructing a different topic vector for each comparison. For instance, we had to exclude volumes bearing the tag ``War fiction'' when ``Love stories'' were compared to that genre, but then exclude a different set of books when ``Love stories'' were compared to ``Bildungsromans.''

\item We also constructed a third set of genre vectors where the topic vector for each volume was first centered by subtracting the running average vector for its position on the timeline. This allowed us to factor out differences produced purely by chronology, and compare genres in terms of their \textit{difference} from the prevailing norm in each period.

\end{enumerate}

\subsection{Comparing the predictions of supervised models}

So far we have envisioned cultural ``distances'' as comparisons between vectors in a single shared space. The angle between ``Bildungsromans'' and ``Short stories'' may differ from the angle between ``Bildungsromans'' and ``Juvenile fiction,'' but both comparisons are made in a space defined by the same set of features (whether topics or words). However, the assumption of a shared space might not be justified. The features that distinguish novels from short stories could simply stop mattering when we turn to the orthogonal contrast between adult and juvenile audiences. One way to acknowledge this is to train a different supervised model for each genre. By assigning different weights to features, supervised models acknowledge that distances are measured, not just in different directions, but in different spaces.

Predictive models could be used to measure distance in several ways. One approach would train a separate model to distinguish each pair of categories, and interpret accuracy (or AUC) as a form of distance. But this approach has several disadvantages. Accuracy and social differentiation seem unlikely to have a linear relationship: it could be much harder to go from 90\% to 95\% accuracy than from 50\% to 55\%. Also, a direct comparison between categories would make it difficult to factor out confounding variables, as we factored out the time axis in 3.2.3 above.

\begin{figure*}
  \includegraphics[width=\textwidth]{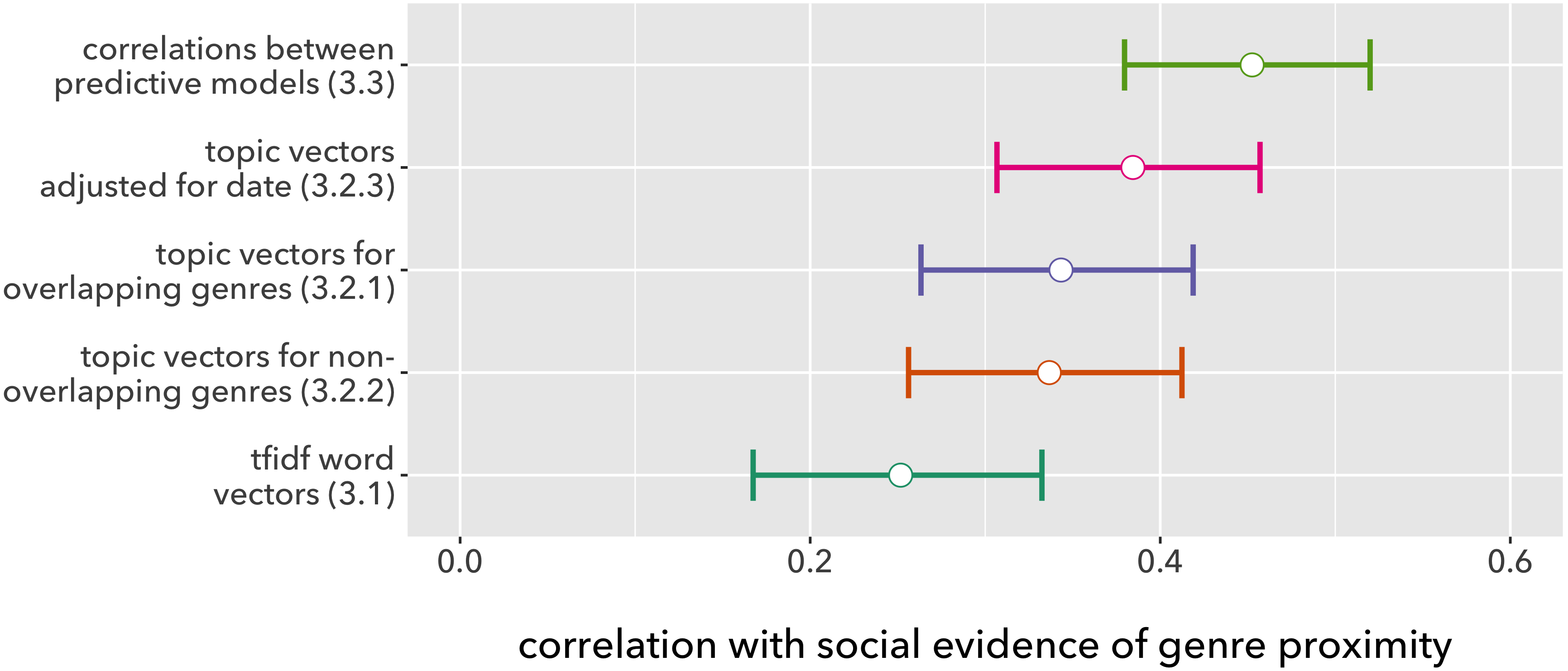}
  \caption{Pearson correlation between textual similarity and evidence of social proximity. Error bars cover a 95\% confidence interval~\cite{zuo:ci,ggplot}.}
  \label{fig:boxplots1}
\end{figure*}

So we chose instead to ``center'' all models on a random sample of fiction with the same distribution across time as the genre category being modeled. Model A (of genre $a$) learns to distinguish short stories from a random sample of fiction; model B (of genre $b$) learns to distinguish juvenile fiction from another random sample (with a different chronological distribution). The number of features and regularization constant for regularized logistic regression are tuned by grid search. Then we apply each model to the works used in the other, and compare the way they rank the books by probability of belonging to $a$ or $b$. This produces two Spearman correlation coefficients; before they can be averaged, they need to undergo Fisher's z-transformation---in effect, $arctanh$. So the distance $d$ between $a$ and $b$ is
\begin{multline}
d(a; b) = -\frac{1}{2}(arctanh(spearman(p(a|A), p(a|B))) + arctanh(spearman(p(b|B), p(b|A))))
\end{multline}
Initial experiments on a toy sample revealed that this measure has a linear relationship to random dilution of data. So Fisher's z-transformation is not just a convenience for averaging; it gives correlation coefficients an unbounded distribution that may be appropriate for a measure of statistical distance. However, other measures might be equally appropriate. For instance, given a well-calibrated probabilistic model, Pearson correlation can be substituted for Spearman in equation (2).

\section{Results}

The measures discussed in Section~\ref{meth} were used to estimate textual distances for all distinct pairings of the 32 categories selected in Section~\ref{data}---except self-comparisons, which were ignored. (Subtracting self-comparisons, $n = 496$.) Different measures of textual distance were compared to the estimates of social distance produced by calculating PMI on genre labels. (Since PMI is actually a measure of proximity, we reversed the sign to interpret it as a distance.) 

The Pearson correlations of different textual measures with our social benchmark are shown in Figure~\ref{fig:boxplots1}. This is admittedly an imperfect evaluative strategy. To understand how these results might confirm or weaken specific methodological priors, details need to be inspected more closely.

\subsection{Social and textual measures of distance do correlate}

Prevailing practice in cultural analytics relies on comparisons between topic vectors~\cite{jockers:macroanalysis,barron:individuals,mauch:evolution}. These measures correlated with social measures of distance at $r = .33$ to $r = .38$, a moderately strong effect size. It is difficult to say exactly how pleased or concerned we should be by correlations around $.35$, because the social evidence itself is sparse and fallible. We don't know that it's a gold standard, and thus don't know how closely textual distances should be approximating it. But it would certainly have been concerning to discover that social and textual measures didn't correlate at all!

\subsection{Overlapping genres did not significantly distort results}

Excluding the intersection of genres---to avoid potential circularity in our comparisons---greatly complicated this study. But in the end, this precaution made little difference. Allowing genres to overlap gave one method (3.2.1) a slight advantage, but the effect was dwarfed by date adjustment or predictive modeling on non-overlapping genres.

\subsection{Topic modeling did make a difference}

Cosine similarity on raw tf-idf vectors performed significantly worse than the other methods. It appears that topic modeling does tend to foreground lexical choices that have social significance. However, this effect might depend as much on corpus construction as it does on the algorithm. In this case, for instance, topics were inferred from a corpus that had been selected to represent generic differences with evenly-sized samples. Many genres would not have been well represented in a purely random sample, and topics inferred from that sample might not have been as well suited to the discrimination of genres.

Note also that a goal of even coverage may not entirely resolve this problem: some empirical evidence suggests that topic models covering a long timeline tend to give a finer-grained description of documents toward the center of the timeline, exaggerating distances there. If this turns out to be a general pattern, it would be a significant problem for diachronic research, deserving further investigation.

\begin{figure}[t!]
\centering
\begin{minipage}{.47\textwidth}
  \centering
  \includegraphics[width=\textwidth]{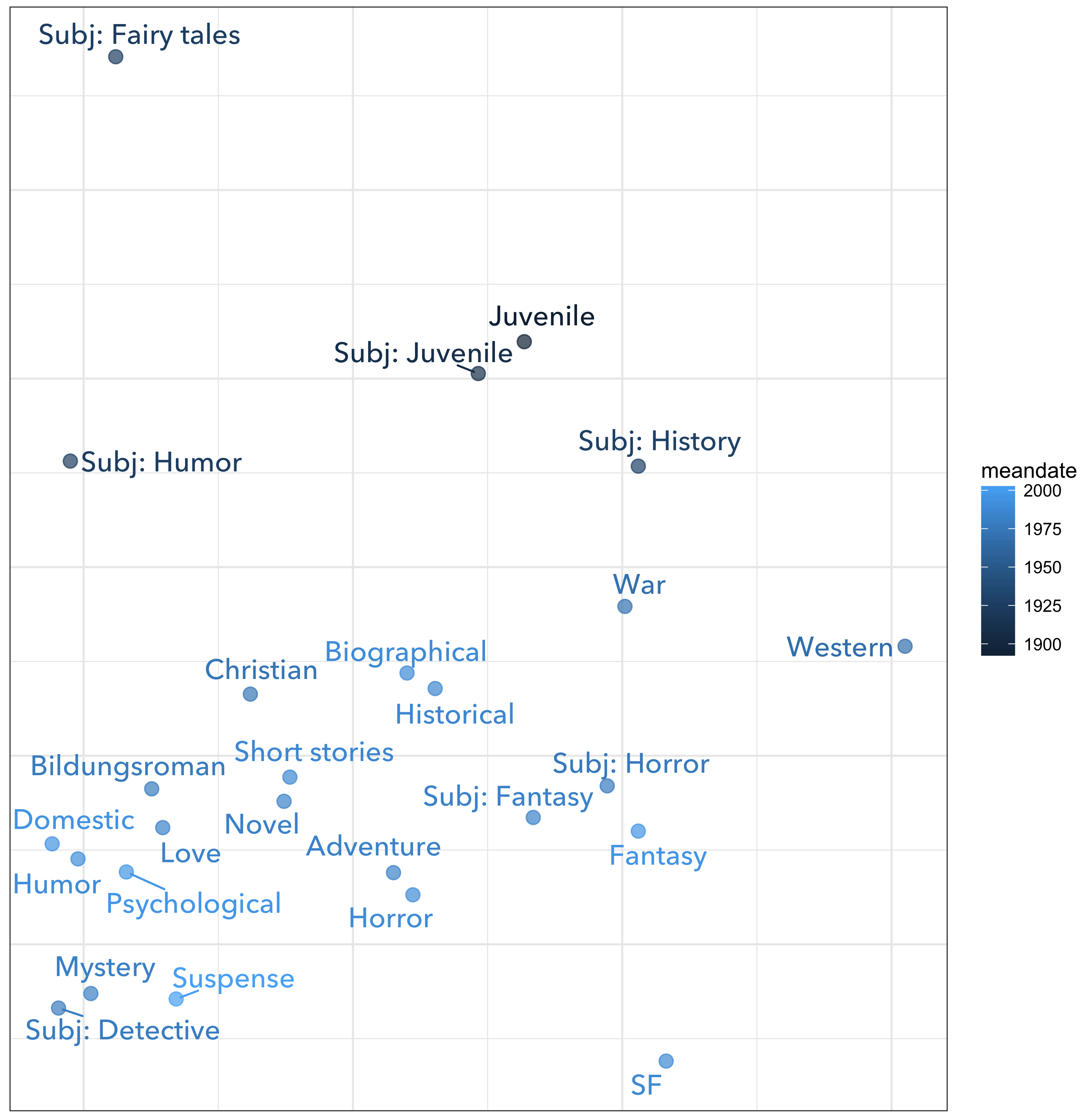}
  \caption{Multidimensional scaling on distances inferred from topic vectors (3.2.2).\\Genres colored by mean publication date.}
  \label{fig:topicmap2}
\end{minipage}%
\hfill
\begin{minipage}{.47\textwidth}
  \includegraphics[width=\textwidth]{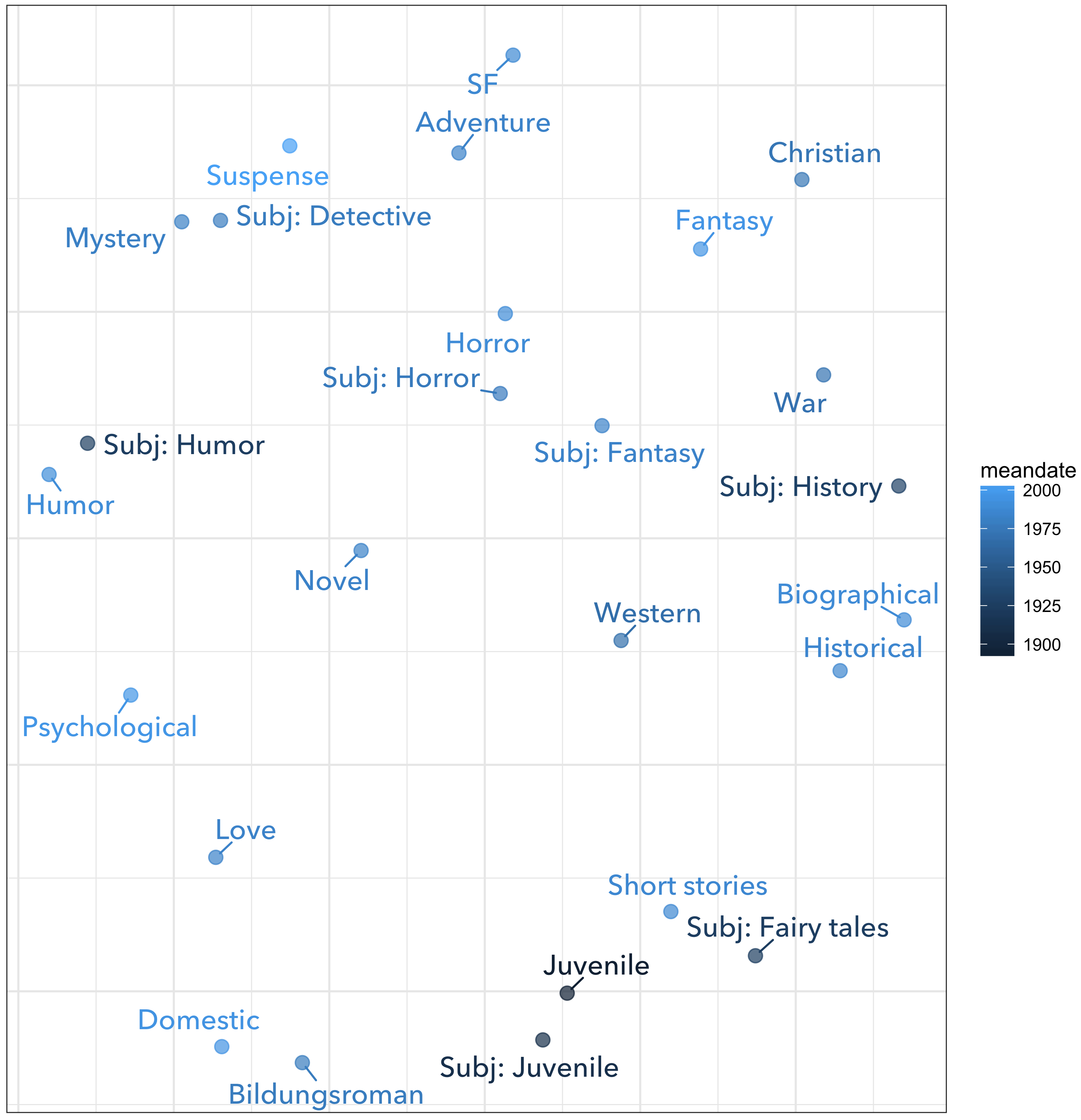}
  \caption{Multidimensional scaling on distances inferred from predictive models (3.3). Genres colored by mean publication date.}
  \label{fig:predictivemap3}
\end{minipage}
\end{figure}

\subsection{Limiting the form of distance being measured is vital for historical significance}

The question posed in this experiment concerned differences between genres. We couldn't assume that those differences were entirely independent of time; e.g. Westerns came along almost a century after historical novels, so the difference between those categories may be partly chronological. But we had reason to suspect that transformation of the English language would often be a confounding variable, obscuring the textual similarities between categories with generic kinship. 

For instance, the subject category ``Humor'' contains volumes with a mean publication date of 1925; it's one of the earlier categories applied to fiction. But the genre category ``Humor'' has a mean date of 1990; as we mentioned in Section~\ref{data}, genres don't get described \textit{as genres} until relatively late in the history of cataloging. The mere passage of time will create many linguistic differences between these groups of books, yet our priors would probably place the categories relatively close to each other.

As you can see on the left side of Figure~\ref{fig:topicmap2}, distances inferred from topic vectors (3.2.2) don't succeed in placing ``Humor'' and ``Subj: Humor'' next to each other. Instead, this space looks suspiciously dominated by a chronological gradient.

By contrast, distances inferred from predictive models place ``Humor'' close to ``Subj: Humor'' and ``Historical'' novels close to ``Subj: History.'' Because our predictive models always use a random contrast set that matches the chronological distribution of the genre category being modeled, the confounding variable of time is partly factored out of distance measurements. Something similar can be achieved by adjusting topic vectors for time (3.2.3).

Note, however, that this choice doesn't compel us to ignore chronological differences. Some genres do change more across time than others. In Figure~\ref{fig:predictivemap3} you may notice that ``Fantasy'' and ``Subj: Fantasy'' are relatively remote, compared to, say, ``Mystery'' and ``Subj: Detective.'' Existing literary scholarship already suggests that the detective story crystallized earlier than the modern fantasy genre, and remained more stable across the last two centuries~\cite{rachman:poe}. 19th-century fantasy, for instance, can be difficult to distinguish from children's literature~\cite{levy:fantasy}. When a genre changes rapidly, different samples will differ from a random contrast set in different ways, even though the contrast sets always match the chronological distribution of the genre. Genres that change rapidly should thus be expected to cover a lot of space in our map.

\begin{wrapfigure}{L}{0.5\textwidth}
  \centering
  \includegraphics[width=.47\textwidth]{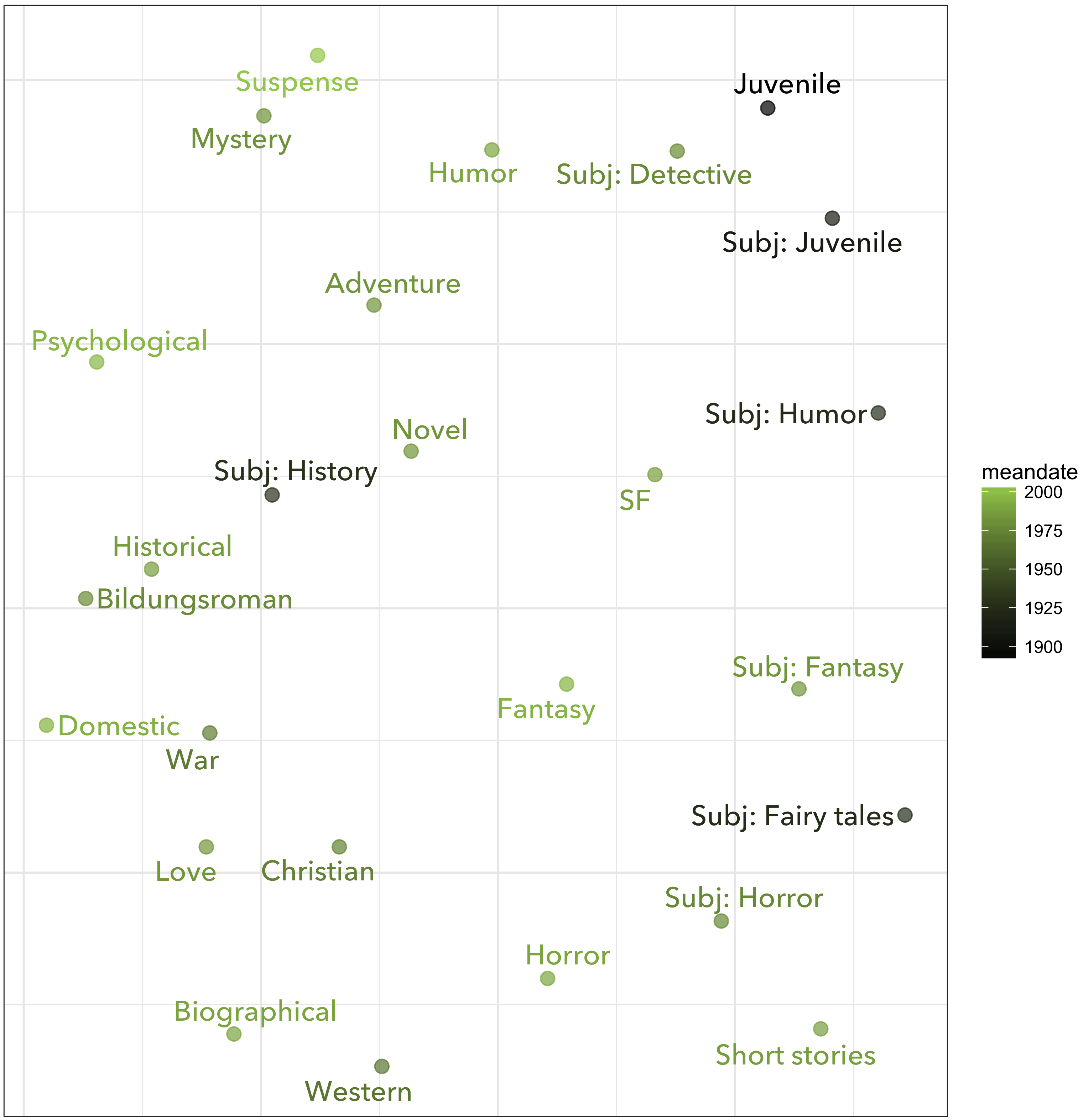}
  \caption{\label{fig:socialmap}Multidimensional scaling on distances inferred from PMI on labels assigned by librarians.}
\end{wrapfigure}

This sort of detailed inspection is an essential supplement to the quantitative evaluation in Figure~\ref{fig:boxplots1}. Social distances inferred from library metadata are perhaps best understood as an initial sanity check; if those measures of social contiguity didn't correlate with textual distances at all, we would need to reconsider this whole research program. But it is not safe to assume that any sample of social evidence constitutes firm ground truth about the real similarities of genres.

In fact, readers might also want to skeptically inspect the social evidence we have been using as a benchmark for textual distances. Figure~\ref{fig:socialmap} makes this possible. Casual inspection is vulnerable to confirmation bias, but in the present author's judgment this map is clearly preferable to Figure~\ref{fig:topicmap2}. For instance, ``Novel'' and ``Short stories'' ought to be fairly remote categories; Figure~\ref{fig:topicmap2} put them too near each other. On the other hand, it is far from clear that Figure~\ref{fig:socialmap} fits our priors better than Figure~\ref{fig:predictivemap3}. Categories like ``Mystery'' and ``Subj: Detective'' ought to be closer than they appear in the social map.

Ironically, Figure~\ref{fig:socialmap} is based on priors specifying, for instance, that ``Mystery'' and ``Subj: Detective'' should be relatively close. But they were only priors, and therefore had to compromise with the evidence provided by library metadata. Figure~\ref{fig:predictivemap3}, which relies purely on textual evidence and not on our priors, actually conforms to those priors more closely. Moreover, it provides surprises that retrospectively make sense (e.g. ``Bildungsroman'' is close to ``Juvenile'' fiction.) Evidence of this kind tends to demonstrate that predictive models provide reliable guidance about the relationships between genres. One can have roughly the same amount of confidence in time-adjusted topic vectors (the method described in 3.2.3). The value of the other three measures of textual distance is more dubious. On inspection, it seems likely that those distance matrices are dominated by a chronological gradient that doesn't correspond to readers' intuitions about the relationships between categories. In other words, the gap between the top two methods in Figure~\ref{fig:boxplots1}, and the bottom three methods, may be bigger than it appears.

Problems of this kind are fairly common. Researchers inquiring about textual distance often tacitly expect to measure certain \textit{kinds} of distance, and tacitly expect other kinds to be factored out. Measurement will become more reliable if those tacit assumptions are made explicit. In this case, time was the problem---and time is definitely a common problem in historical inquiry. But other research projects, inquiring about the pace of stylistic change for instance, may confront a converse problem. Time might become the topic of interest, and the distribution of texts across genres might become the confounding variable. The solutions explored here would still be applicable. For instance, to factor generic differences out of a question about time, one wouldn't necessarily have to hold genre distribution constant across the timeline (and risk distorting the corpus). Instead, one could train predictive models where the genre distribution of the random contrast set matches the genre distribution of the decade modeled. That strategy can reduce confounds while still acknowledging that genre and time are interwoven.

\section{Conclusions}

This paper asks whether the assumptions underpinning contemporary work in cultural analytics hold up to empirical scrutiny. Broadly, it has answered ``yes.'' Textual distances do correlate with independent measures of social affinity. And topic modeling a corpus probably does improve that correlation.

On the other hand, there are also reasons for caution here. Figure~\ref{fig:topicmap2} suggests that topic modeling cannot in itself replace reflection on confounding variables. Unless researchers explicitly limit the range of differences that interest them, general measurements of ``stylistic change'' may indiscriminately register linguistic drift, the rise of a new genre, or a shift in the underlying gender balance of authorship.

This paper has not prescribed a particular response to the problem; two different methods performed well in this experiment, and other solutions are imaginable. However, the strong performance of supervised models ($r = .45$) does hint that they have under-exploited potential as measures of distance. 

\begin{wrapfigure}{L}{0.48\textwidth}
  \centering
  \includegraphics[width=.44\textwidth]{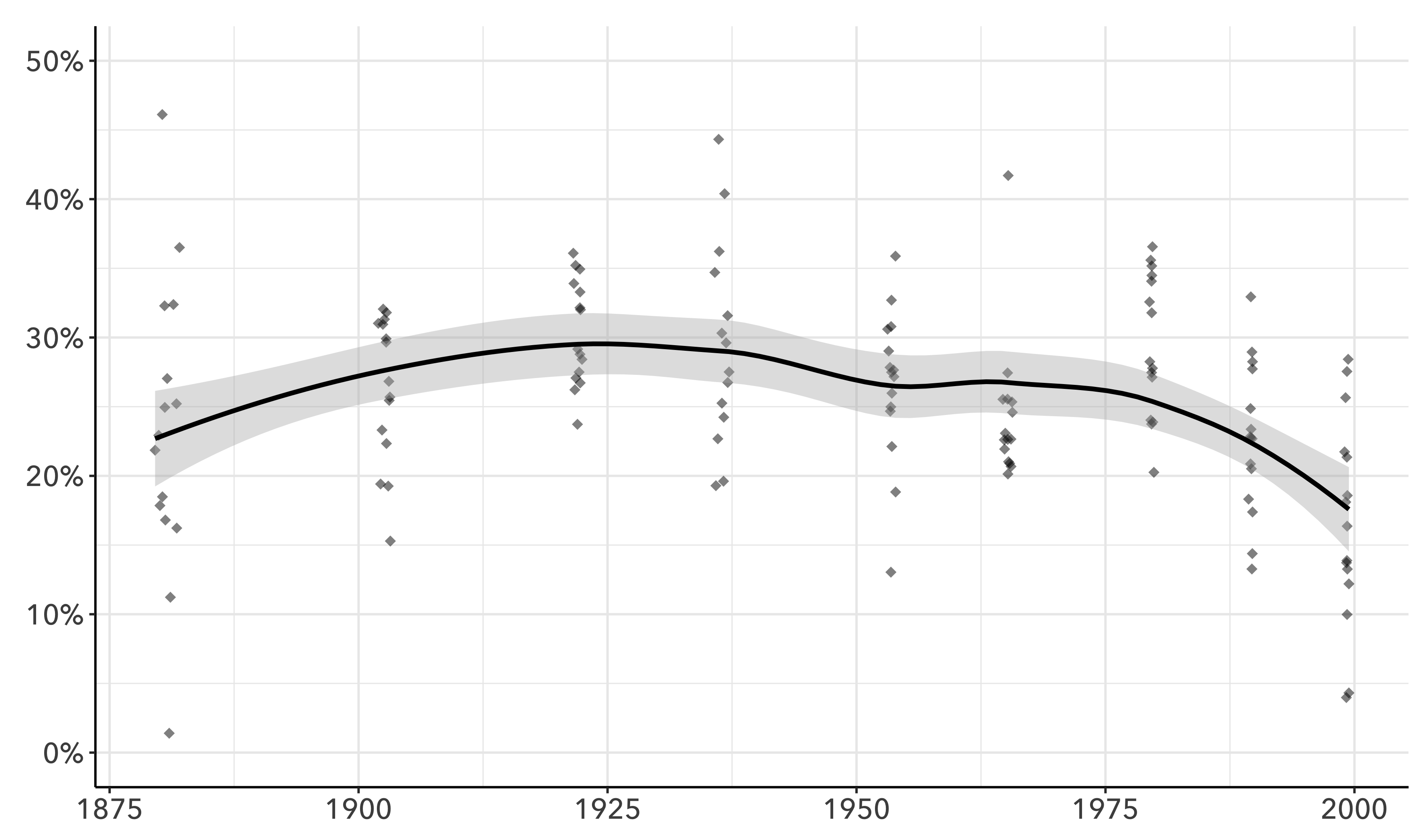}
  \caption{\label{fig:fsf}Distance between models of fantasy and science fiction, expressed as a percentage of the text that would need to be randomly altered to produce the measured loss of correlation.}
\end{wrapfigure}

Supervised learning is usually explained as if it aimed simply to reproduce existing ground truth. If this were true, supervised models would find little application in the humanities. Humanists are often posing questions where reliable ground truth is not yet available (and perhaps never will be). ``Accuracy'' would be a na\"ive goal.

However, predictive models don't need to be contained in that na\"ive frame. In this paper, for instance, models of science fiction were applied to fantasy---not because we expected them to be right, but to measure divergent assumptions. Training a predictive model can translate a static set of examples into a measure of distance, and thereby put concepts in dialogue with each other.

Our goal in this experiment was simply to develop a measure of the distance between genres that corresponds to literary intuition, because it uses evidence about the genres themselves to define ``distance.'' But once we have done that, it becomes possible to pose questions where we don't have intuitive answers. Figure 5, borrowed from a forthcoming project, hints that the genres of science fiction and fantasy have been converging in the last thirty years. Predictive models can also give scholars a way of comparing different perspectives on genre. Instead of taking library metadata as authoritative, we can draw lists of examples from the conflicting practices of critics or reviewers, and compare the resulting models.

How widely could this method be applied? The history of genre may initially appear to be an odd special case, where we happen to have labeled categories. The methods that work for genre may seem unlikely to be useful for more open-ended questions about historical change. 

But labels come in many forms. Time itself is a label. For instance, one might ask, ``How quickly has science fiction changed over the last century?'' and answer by comparing models of the genre trained on different segments of the timeline.

The task of measuring distance in a supervised way, triangulating from multiple reference points, needs better computational solutions. We readily admit that the solution offered in this paper was improvised empirically and uses Spearman correlation as duct tape. It works, but it has awkward limitations: for instance, to put models on an equal footing, all categories have to be limited to an equal number of examples. A more integrated and principled approach is imaginable, patterned perhaps on supervised topic modeling~\cite{NIPS2008_3599,Zhu09medlda}, but with the goal of inferring distances (of a particular social character) rather than latent topics.

\section{Open code and data}
\label{sec:supplemental}
Code used in this paper is available both on GitHub \url{https://github.com/tedunderwood/genredistance}, and more archivally at Zenodo~\cite{underwood:code}. Data used in the paper has also been archived~\cite{underwood:data}.

\section*{Acknowledgments}

The research undertaken here was supported by the NovelTM project, funded by Canada's Social Sciences and Humanities Research Council and directed by Andrew Piper. HathiTrust Research Center was also a crucial resource, and some work was supported by the Andrew W. Mellon Foundation via the WCSA+DC grant to HTRC, directed by J. Stephen Downie. At an early stage of the project, crucial advice came from Jacob Eisenstein; conversation with David Mimno and Laure Thompson guided some of my thinking about topic models.

\bibliographystyle{acl}
\bibliography{UnderwoodACL2018}

\begin{thebibliography}{}

\bibitem[\protect\citename{Barron \bgroup et al.\egroup
  }2018]{barron:individuals}
Alexander T.~J. Barron, Jenny Huang, Rebecca~L. Spang, and Simon DeDeo.
\newblock 2018.
\newblock Individuals, institutions, and innovation in the debates of the
  {F}rench {R}evolution.
\newblock {\em Proceedings of the National Academy of Sciences}.

\bibitem[\protect\citename{Biber and Conrad}2009]{biberconrad:genre}
Douglas Biber and Susan Conrad.
\newblock 2009.
\newblock {\em Register, Genre, and Style}.
\newblock Cambridge University Press, Cambridge.

\bibitem[\protect\citename{Capitanu \bgroup et al.\egroup }2016]{capitanu}
Boris Capitanu, Ted Underwood, Peter Organisciak, Timothy Cole, Maria~Janina
  Sarol, and J.~Stephen Downie.
\newblock 2016.
\newblock The {H}athi{T}rust {R}esearch {C}enter extracted feature dataset
  (1.0).

\bibitem[\protect\citename{Cha}2007]{cha2007comprehensive}
Sung-Hyuk Cha.
\newblock 2007.
\newblock Comprehensive survey on distance/similarity measures between
  probability density functions.
\newblock {\em International Journal of Mathematical Models and Methods in
  Applied Sciences}, 1(4):300--307.

\bibitem[\protect\citename{Di{M}aggio}1987]{dimaggio:genre}
Paul Di{M}aggio.
\newblock 1987.
\newblock Classification in art.
\newblock {\em American Sociological Review}, 52:440--455.

\bibitem[\protect\citename{Jockers}2013]{jockers:macroanalysis}
Matthew~L. Jockers.
\newblock 2013.
\newblock {\em Macroanalysis: Digital Methods and Literary History}.
\newblock University of Illinois Press, Urbana.

\bibitem[\protect\citename{Jones}1972]{sparck:tfidf}
Karen~Spärck Jones.
\newblock 1972.
\newblock A statistical interpretation of term specificity and its application
  in retrieval.
\newblock {\em Journal of Documentation}, 28(1):11--21.

\bibitem[\protect\citename{Kessler \bgroup et al.\egroup
  }1997]{kessler:automatic}
Brett Kessler, Geoffrey Nunberg, and Hinrich Schütze.
\newblock 1997.
\newblock Automatic detection of text genre.
\newblock In {\em Proceedings of the Eighth Conference on European Chapter of
  the Association for Computational Linguistics}, EACL '97, pages 32--38,
  Stroudsburg, PA, USA. Association for Computational Linguistics.

\bibitem[\protect\citename{Kim \bgroup et al.\egroup }2017]{kim:genre}
Evgeny Kim, Sebastian Padó, and Roman Klinger.
\newblock 2017.
\newblock Investigating the relationship between literary genres and emotional
  plot development.
\newblock In {\em Proceedings of the Joint SIGHUM Workshop on Computational
  Linguistics for Cultural Heritage, Social Sciences, Humanities, and
  Literature}, pages 17--26. Association for Computational Linguistics.

\bibitem[\protect\citename{Lacoste-Julien \bgroup et al.\egroup
  }2009]{NIPS2008_3599}
Simon Lacoste-Julien, Fei Sha, and Michael~I. Jordan.
\newblock 2009.
\newblock Disc{LDA}: Discriminative learning for dimensionality reduction and
  classification.
\newblock In D.~Koller, D.~Schuurmans, Y.~Bengio, and L.~Bottou, editors, {\em
  Advances in Neural Information Processing Systems 21}, pages 897--904. Curran
  Associates, Inc.

\bibitem[\protect\citename{Levy and Mendlesohn}2016]{levy:fantasy}
Michael Levy and Farah Mendlesohn.
\newblock 2016.
\newblock {\em Children's Fantasy Literature: An Introduction}.
\newblock Cambridge University Press, Cambridge.

\bibitem[\protect\citename{Mauch \bgroup et al.\egroup }2015]{mauch:evolution}
Matthias Mauch, Robert~M. MacCallum, Mark Levy, and Armand~M. Leroi.
\newblock 2015.
\newblock The evolution of popular music: {U}{S}{A} 1960{\textendash}2010.
\newblock {\em Royal Society Open Science}, 2(5).

\bibitem[\protect\citename{Miller}2000]{miller:genre}
David~P. Miller.
\newblock 2000.
\newblock Out from under: Form/genre access in {L}{C}{S}{H}.
\newblock {\em Cataloging and Classification Quarterly}, 29(1/2):169--188.

\bibitem[\protect\citename{Pedregosa \bgroup et al.\egroup }2011]{scikit-learn}
F.~Pedregosa, G.~Varoquaux, A.~Gramfort, V.~Michel, B.~Thirion, O.~Grisel,
  M.~Blondel, P.~Prettenhofer, R.~Weiss, V.~Dubourg, J.~Vanderplas, A.~Passos,
  D.~Cournapeau, M.~Brucher, M.~Perrot, and E.~Duchesnay.
\newblock 2011.
\newblock Scikit-learn: Machine learning in {P}ython.
\newblock {\em Journal of Machine Learning Research}, 12:2825--2830.

\bibitem[\protect\citename{Rachman}2010]{rachman:poe}
Stephen Rachman.
\newblock 2010.
\newblock Poe and the origins of detective fiction.
\newblock In Catherine~Ross Nickerson, editor, {\em The Cambridge Companion to
  American Crime Fiction}, pages 17--28. Cambridge University Press, Cambridge.

\bibitem[\protect\citename{Stone}2000]{stone:lcsh}
Alva~T. Stone.
\newblock 2000.
\newblock The {L}{C}{S}{H} century: A brief history of the {L}ibrary of
  {C}ongress subject headings, and introduction to the centennial essays.
\newblock {\em Cataloging and Classification Quarterly}, 29(1/2):1--15.

\bibitem[\protect\citename{Underwood}2018a]{underwood:code}
Ted Underwood.
\newblock 2018a.
\newblock Code to support "{T}he historical significance of textual distances".
\newblock \url{https://zenodo.org/record/1300934#.WzaQvSOZNBw}.

\bibitem[\protect\citename{Underwood}2018b]{underwood:data}
Ted Underwood.
\newblock 2018b.
\newblock Data on the historical significance of textual distances.
\newblock \url{http://hdl.handle.net/2142/100119}.

\bibitem[\protect\citename{Wickham}2009]{ggplot}
Hadley Wickham.
\newblock 2009.
\newblock {\em ggplot2: Elegant Graphics for Data Analysis}.
\newblock Springer-Verlag New York.

\bibitem[\protect\citename{Wu \bgroup et al.\egroup }2010]{wu:genre}
Zhili Wu, Katja Markert, and Serge Sharoff.
\newblock 2010.
\newblock Fine-grained genre classification using structural learning
  algorithms.
\newblock In {\em Proceedings of the 48th Annual Meeting of the Association for
  Computational Linguistics}, pages 749--759. Association for Computational
  Linguistics.

\bibitem[\protect\citename{Zhu \bgroup et al.\egroup }2009]{Zhu09medlda}
Jun Zhu, Amr Ahmed, Eric~P. Xing, Jun Zhu, Amr Ahmed, Eric~P. Xing, and David
  Blei.
\newblock 2009.
\newblock Med{LDA}: Maximum margin supervised topic models for regression and
  classification.
\newblock In {\em In Leon Bottou and Michael Littman, editors, International
  Conference on Machine Learning (ICML)}, pages 1257--1264.

\bibitem[\protect\citename{Zuo}2018]{zuo:ci}
Zhiya Zuo.
\newblock 2018.
\newblock Calculate {P}earson correlation interval in python.
\newblock \url{https://zhiyzuo.github.io/Pearson-Correlation-CI-in-Python/}.

\end{thebibliography}

\end{document}